%% file: main.tex
\documentclass[10pt,twocolumn,letterpaper]{article}

\usepackage{cvpr}              
\usepackage{url}

\usepackage{graphicx}
\usepackage{algorithmic}
\usepackage{algorithm}

\usepackage{arydshln}
\usepackage{multirow}
\usepackage{booktabs}

\usepackage{multirow}
\usepackage{booktabs}
\usepackage{makecell}
\usepackage{enumitem}
\usepackage{makecell}

\input{preamble}
\definecolor{cvprblue}{rgb}{0.21,0.49,0.74}
\usepackage[pagebackref,breaklinks,colorlinks,allcolors=cvprblue]{hyperref}

\def\paperID{8788} 
\def\confName{CVPR}
\def\confYear{2026}

\title{Mitty: Diffusion-based Human-to-Robot Video Generation}

\author{
  Yiren Song \quad
  Cheng Liu \quad
  Weijia Mao \quad
  Mike Zheng Shou\textsuperscript{$\dagger$} \\
  \textsuperscript{1}Show Lab, National University of Singapore \\
}

\begin{document}

\twocolumn[{%
\renewcommand\twocolumn[1][]{#1}%
\maketitle

\begin{center}
   \captionsetup{type=figure}

\includegraphics[width=\linewidth]{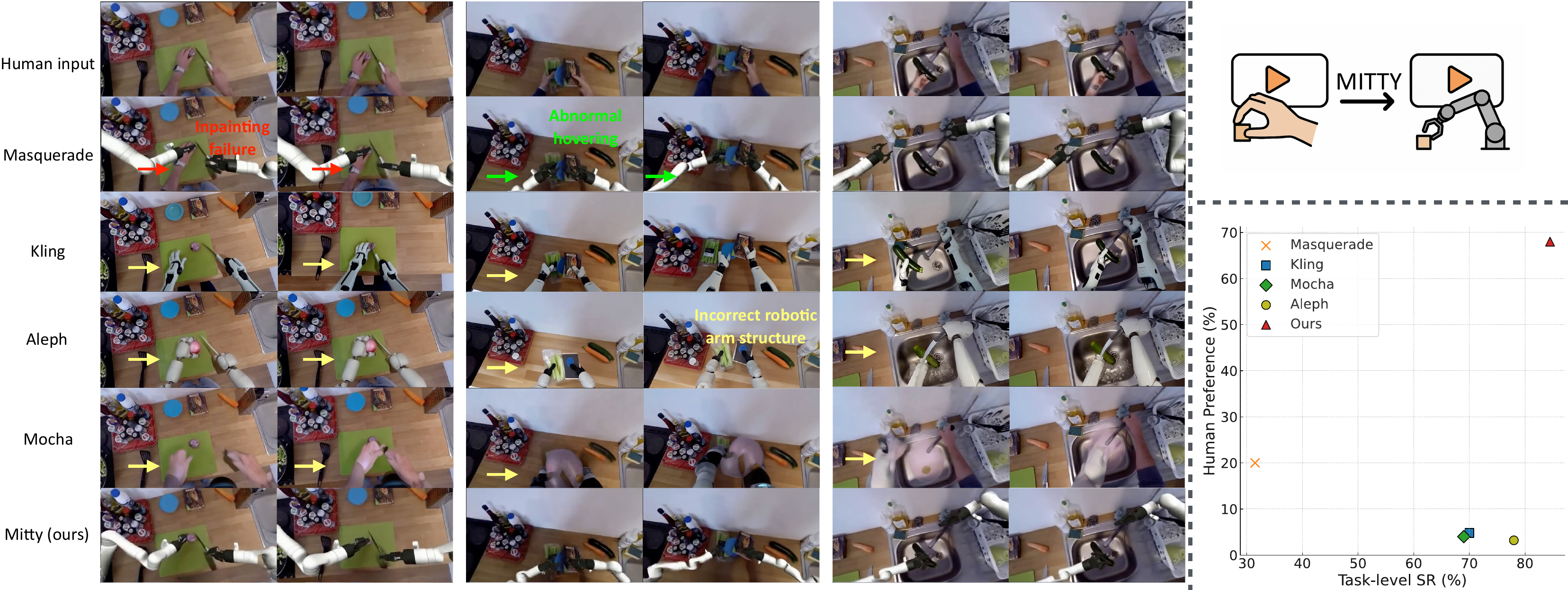}    
    \setlength{\abovecaptionskip}{0pt}
    \caption{We propose Mitty, a paradoxical in-context learning–based video generation method built on Diffusion Transformers. It can convert human demonstration videos into robotic manipulation videos and achieves high task success rates. 
}
    \label{teaser} 

\end{center}   
}]

\begingroup
\renewcommand\thefootnote{}
\footnotetext{$\dagger$ Corresponding author.}
\endgroup

\begin{abstract}

Learning directly from human demonstration videos is a key milestone toward scalable and generalizable robot learning. Yet existing methods rely on intermediate representations such as keypoints or trajectories, introducing information loss and cumulative errors that harm temporal and visual consistency. We present Mitty, a Diffusion Transformer that enables video In-Context Learning for end-to-end Human2Robot video generation. Built on a pretrained video diffusion model, Mitty leverages strong visual-temporal priors to translate human demonstrations into robot-execution videos without action labels or intermediate abstractions. Demonstration videos are compressed into condition tokens and fused with robot denoising tokens through bidirectional attention during diffusion. To mitigate paired-data scarcity, we also develop an automatic synthesis pipeline that produces high-quality human-robot pairs from large egocentric datasets. Experiments on Human2Robot and EPIC-Kitchens show that Mitty delivers state-of-the-art results, strong generalization to unseen environments, and new insights for scalable robot learning from human observations. Our project homepage is available at
\href{https://github.com/showlab/Mitty}{https://github.com/showlab/Mitty}
\end{abstract}

\section{Introduction}

Humans excel at rapidly acquiring new skills by observing others. If robots could directly learn manipulation policies from a single human demonstration video and generate corresponding robot-execution videos, this would provide a critical path toward cross-task and cross-environment generalization. Yet achieving this has long been a highly challenging goal in robotics.

Existing approaches typically rely on intermediate representations such as keypoints, trajectories, or depth maps to bridge human and robot videos. They first extract keypoints or trajectories from the human demonstration and then condition a rendering module to synthesize robot execution videos. While intuitive, this approach fails to fully exploit the rich information embedded in demonstration videos and struggles to capture the fine-grained spatio-temporal dynamics essential for robust generalization. Moreover, errors accumulated in the intermediate estimation stage can further degrade performance. This raises a natural question: can we bypass intermediate representations and directly achieve end-to-end Human2Robot video generation?

This task presents several key challenges: (1) Appearance and scene consistency—the generated robot video must match the scene of the human demonstration while preserving a stable, plausible robot embodiment;  (2) Action and strategy alignment—the robot’s actions must follow the human demonstration yet adapt to structural differences between human hands and robot arms; (3) Data scarcity—despite abundant human and robot videos separately, finely aligned human–robot video pairs are extremely rare. The only public H2R dataset currently contains just 2,600 video pairs across nine tasks, making it difficult to learn generalizable skills from limited data.

To alleviate this scarcity, we employ an automatic paired-data synthesis pipeline \cite{lepert2025masquerade} using egocentric human videos. Starting from large-scale human activity datasets such as EPIC-Kitchens, we estimate 3D hand keypoints, remove human hands, and inpaint clean backgrounds. We then map keypoint sequences to robot end-effector poses and render robot arms into the video, producing high-quality human–robot paired videos. This approach bypasses traditional intermediate representations and significantly improves both scale and fine-grained temporal consistency, providing stronger training and generalization capacity for our model.

Building on this foundation, we introduce Mitty, a Diffusion Transformer framework for video In-Context Learning. In-context learning (ICL) has shown promise for fewshot learning, offering data-efficient and rapid adaptation at test time. By simply conditioning on one human demonstrations, ICL can predict robot actions to achieve novel tasks at test time without expensive retraining. Our method conditions directly on human demonstration videos to generate corresponding robot-execution videos in an end-to-end manner, requiring no explicit action labels. Mitty leverages Wan 2.2, a powerful video generation model pretrained on massive natural video corpora, to inherit strong visual and temporal priors. Concretely, we compress the human demonstration into condition tokens via a VAE (kept noise-free) and concatenate them with the robot denoise tokens through a bidirectional attention mechanism during diffusion, enabling cross-domain action translation. Mitty supports two inference modes—first-frame-controlled and zero-frame generation—offering greater flexibility in deployment. We further conduct a systematic evaluation across models and settings to provide actionable insights for the community.

Across both the Human2Robot dataset and EPIC-Kitchens, Mitty significantly outperforms existing baselines and demonstrates strong generalization. Although our work does not yet realize a full Human2Robot control loop, generating high-fidelity, temporally aligned, and semantically accurate Human2Robot videos is itself a crucial and challenging task. Such videos offer richer supervision than keypoints or trajectories and form the foundation for future video-to-policy inversion. Once the generated videos are reliable enough, they hold the potential to be translated into executable control signals, making high-quality Human2Robot video generation a key first step toward end-to-end Human2Robot mapping. We summarize our contributions as follows:

\begin{enumerate}[leftmargin=*]
    \item We propose Mitty, the first end-to-end Human2Robot video generation framework built upon a Video Diffusion Transformer.
    \item Technically, we leverage in-context learning to achieve both appearance and scene consistency as well as action consistency, significantly improving cross-task generalization.
    \item We design an efficient data synthesis strategy and combine it with existing datasets for mixed training, which markedly enhances the model’s generalization ability on unseen tasks and environments. Extensive experiments demonstrate the effectiveness and superiority of our approach in terms of generation quality and cross-task consistency.
\end{enumerate}

\section{Related Works}
\subsection{Video Generation Models}
Video generation models have evolved rapidly from early GAN-based approaches \cite{pan2017create}, UNet-based approaches \cite{guo2023animatediff, xu2024magicanimate, song2024processpainter} to today’s Diffusion Transformer architectures \cite{peebles2023scalable, wan2025wan, zheng2024open, jiang2025vace}. Modern Diffusion Transformers can generate high-quality, temporally coherent videos conditioned on text, images, or multi-modal inputs, enabling applications such as controllable video generation \cite{lin2025omnihuman, jiang2025vace, ma2025followyourclick, ma2025followfaster, ma2025followyourmotion} and world modeling \cite{gao2025adaworld}. Many recent studies also leverage large pretrained video generation models for tasks in robotics and mechanical manipulation \cite{fu2025learning}, highlighting their potential for cross-domain generalization and interactive learning.

\subsection{Learning from Human Videos}
A growing body of work investigates how large human-centric video datasets can be used to improve robot policy learning \cite{xie2025human2robot, shah2025mimicdroid, lepert2503phantom, videodex, dexvip, track2act, egomimic, egozero, yang2025x, li2025mimicdreamer, team2025gigaworld}. Compared to costly and time-consuming teleoperation, large-scale human videos provide a scalable and diverse source of demonstrations. Earlier studies focused on extracting visual representations \cite{chen2025vidbot}, deriving reward functions \cite{guzey2025bridging}, or directly estimating motion priors from human videos \cite{wang2023mimicplay, qiu2025humanoid}. However, many approaches still rely on additional robot data or specialized hardware such as VR and hand-tracking devices, limiting scalability. Recent progress in 3D hand pose estimation \cite{cheng2024handdiff} helps extract action information directly from RGB videos, but cross-embodiment transfer remains difficult. Humanoid robots can partially alleviate this gap due to their kinematic similarity to humans. Building on these trends, we propose Mitty, which achieves end-to-end generation of robot videos directly from human demonstrations without extracting intermediate representations such as pose, trajectories, or depth, and better leverages the fine-grained details contained in the original human demonstration videos.

\subsection{In-Context Learning}  
In-context learning (ICL) \cite{brown2020language, alayrac2022flamingo} has demonstrated remarkable capability for adapting models to new tasks at inference time. In the visual generation domain, recent approaches  have leveraged ICL to achieve high-quality image generation \cite{zhang2025context, huang2024context, zhang2025easycontrol, song2025makeanything, song2025layertracer, huang2025photodoodle, song2025omniconsistency, guo2025any2anytryon, wang2025diffdecompose, gong2025relationadapter, lu2025easytext, shi2024fonts, jiang2025personalized} and video generation \cite{zhang2024video, kim2025videoicl, yu2025context}. In robotics, preliminary studies \cite{shah2025mimicdroid} have explored applying ICL to visuomotor policies using either teleoperation or simulation data. However, these methods are constrained by data collection costs and limited task diversity, making large and heterogeneous datasets essential for effective adaptation. We adopt an In-Context Learning framework built on the Wan 2.2 video diffusion model to translate human demonstration videos into robot-arm executions, ensuring visual and action consistency throughout generation.

\section{Method}
In this section, we first define our problem formulation and pverall architecture in Sec.~\ref{sec:problem_architecture}, then we describe in detail how video in-context learning is achieved via bidirectional attention in Sec.~\ref{sec:video_icl}, and finally explain our synthetic paired-data construction pipeline in Sec.~\ref{sec:dataset_construction}.

\subsection{Overall Architecture}
\label{sec:problem_architecture}

We formulate Human2Robot video generation as a conditional denoising problem.  
Given paired data consisting of a human demonstration video $V^H=\{v^H_{1},\dots,v^H_{N}\}$ and the corresponding robot execution video $V^R=\{v^R_{1},\dots,v^R_{N}\}$, our objective is to model the conditional distribution $p_\theta(V^R|V^H)$ that captures fine-grained spatio-temporal correspondences between human actions and robot executions.  
We consider two settings:  
(i) \textbf{H2R}(Human2Robot Video Generation), where the model directly generates a robot execution video from a human demonstration without providing any initial robot frame; and  
(ii) \textbf{HI2R}(Human-and-Initial-Image-to-Robot Video Generation), which extends H2R by additionally supplying an initial robot frame to define the robot’s initial state and guide embodiment and motion planning.

We implement this formulation in a single unified framework built upon Wan 2.2 \cite{wan2025wan}, a state-of-the-art diffusion-based video generation model pretrained on large-scale natural videos. Both human and robot videos are encoded into latent tokens using the same VAE-based video encoder. Human latents act as clean conditioning tokens, while robot latents act as denoising targets. These tokens are concatenated along the temporal dimension and fed into a Diffusion Transformer enhanced with bidirectional attention, enabling information to flow between modalities at each denoising step. This unified design supports both zero-frame generation (H2R) and first-frame-conditioned generation (HI2R), sharing parameters and priors across both settings while allowing fine-grained control over the robot’s initial state and stable motion planning across diverse tasks.

\begin{figure*}[htbp!]
    \centering
    \includegraphics[width=1.0\linewidth]{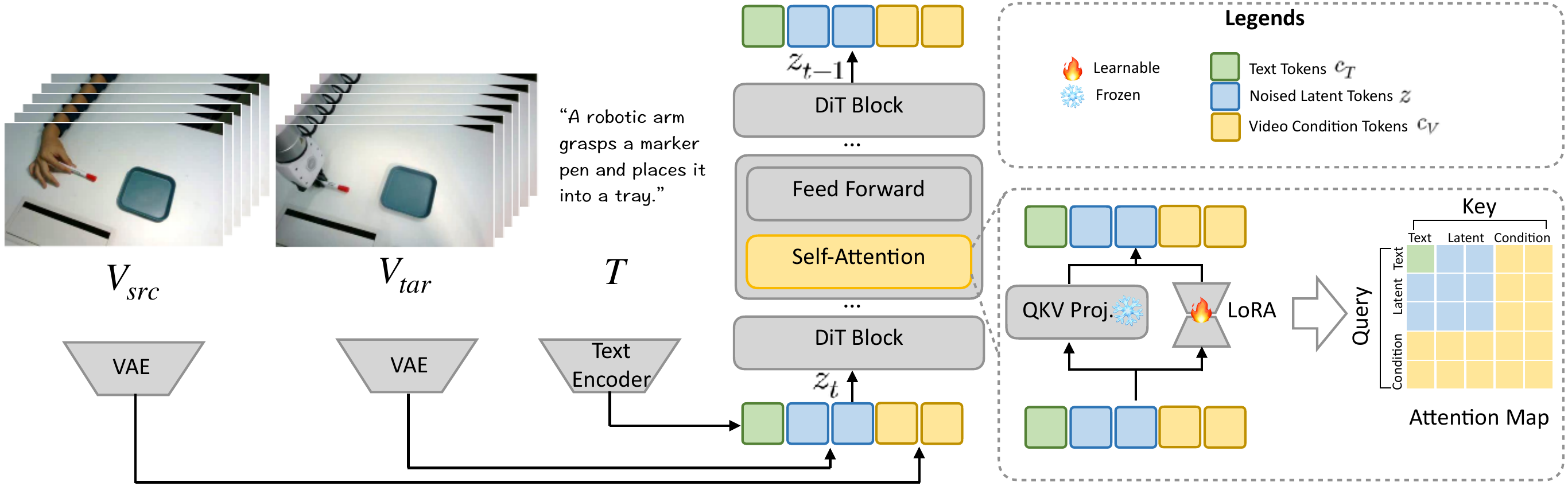}
    \caption{Overall architecture of Mitty. We build Mitty on a Diffusion Transformer–based video generation model and employ an In-Context Learning paradigm. The human demonstration video (input) and the noisy robot video latents (denoise stream) are concatenated, with noise injected only into the robot branch. A bidirectional attention mechanism enables cross-modal information flow, allowing the model to learn to generate robotic videos  directly from human operation demonstrations.}
    \label{fig:results}
\end{figure*}

\subsection{Video In-Context Learning via Bidirectional Attention}
\label{sec:video_icl}

To achieve cross-domain video in-context learning between human and robot modalities, we enhance the Diffusion Transformer with a \textbf{bidirectional attention mechanism} linking human-condition tokens and robot-denoise tokens. This allows the model to dynamically align temporal cues, motion patterns, and object interactions across domains while leveraging the strong visual–temporal priors from the pretrained video backbone.

\textbf{Diffusion Process and Noise Injection.}
Let $\mathbf{z}^R_0=\mathrm{VAE}_{\mathrm{enc}}(\mathbf{V}^R)$ denote the robot video latent.
During training, we progressively add noise only to the robot latents while keeping the human latents clean:
\begin{equation}
\mathbf{x}^R_t=\sqrt{\bar{\alpha}_t}\,\mathbf{z}^R_0+\sqrt{1-\bar{\alpha}_t}\,\boldsymbol{\epsilon},
\quad \boldsymbol{\epsilon}\sim\mathcal{N}(\mathbf{0},\mathbf{I}).
\label{eq:robot_forward_noise}
\end{equation}
with cumulative noise schedule
\begin{equation}
\bar{\alpha}_t=\prod_{s=1}^{t}\alpha_s,\quad t\in\{1,\dots,T\}.
\label{eq:alpha_bar}
\end{equation}
This setup enables us to model the conditional distribution \(p_\theta(\mathbf{V}^R\!\mid \mathbf{V}^H)\) without requiring explicit action or trajectory labels.

\textbf{Token Representation and Embeddings.}
Let $\mathbf{z}^H_0=\mathrm{VAE}_{\mathrm{enc}}(\mathbf{V}^H)$ denote the human video latent. Tokens are formed as
\begin{equation}
\begin{aligned}
\mathbf{C} &= \mathbf{z}^H_0+\mathbf{E}_{\text{time}}+\mathbf{E}_{\text{mod}(h)},\\
\mathbf{D} &= \mathbf{x}^R_t+\mathbf{E}_{\text{time}}+\mathbf{E}_{\text{mod}(r)}.
\end{aligned}
\label{eq:tokens_CD}
\end{equation}
Here $d$ denotes the token/channel dimension and $\mathbf{E}_{\text{time}}$, $\mathbf{E}_{\text{mod}(\cdot)}$ are temporal and modality embeddings.

\textbf{Bidirectional Attention Coupling.}
At each layer, we exchange information in both directions (row-wise softmax):
\begin{equation}
\begin{aligned}
\tilde{\mathbf{C}} &= 
\mathrm{Softmax}\!\Bigl(\tfrac{\mathbf{C}\mathbf{D}^{\top}}{\sqrt{d}}\Bigr)\mathbf{D},\\
\tilde{\mathbf{D}} &= 
\mathrm{Softmax}\!\Bigl(\tfrac{\mathbf{D}\mathbf{C}^{\top}}{\sqrt{d}}\Bigr)\mathbf{C}.
\end{aligned}
\label{eq:attn_CD}
\end{equation}
The updated tokens $[\tilde{\mathbf{C}};\tilde{\mathbf{D}}]$ are concatenated along the token
dimension and fed to subsequent Transformer blocks.

\textbf{Denoising and Reverse Update.}
The network predicts $\boldsymbol{\epsilon}_\theta(\mathbf{x}^R_t,\mathbf{C},t)$ on the robot branch and performs
\begin{equation}
\begin{gathered}
\mathbf{x}^R_{t-1}
=\frac{1}{\sqrt{\alpha_t}}\!\left(
\mathbf{x}^R_t-\frac{1-\alpha_t}{\sqrt{1-\bar{\alpha}_t}}\,
\boldsymbol{\epsilon}_\theta(\mathbf{x}^R_t,\mathbf{C},t)
\right)
+\sigma_t\mathbf{z}, \\
\mathbf{z} \sim \mathcal{N}(\mathbf{0},\mathbf{I}),
\end{gathered}
\label{eq:reverse_update}
\end{equation}

with $\sigma_t$ determined by the variance schedule. The final video is


\begingroup
\setlength\abovedisplayskip{4pt}
\setlength\belowdisplayskip{4pt}
\begin{equation}
\hat{\mathbf{V}}^R=\mathrm{VAE}_{\mathrm{dec}}(\mathbf{z}^R_0).
\end{equation}
\endgroup

This models the conditional generation without action labels, and supports both
\emph{H2R (Zero-frame)} and \emph{HI2R (First-frame conditioned)} modes defined in Sec.~\ref{sec:problem_architecture}, enabling either generation from human demonstrations or controlled execution with an initial robot frame.

\begin{figure*}[!t]
    \centering
    \includegraphics[width=1.0\linewidth]{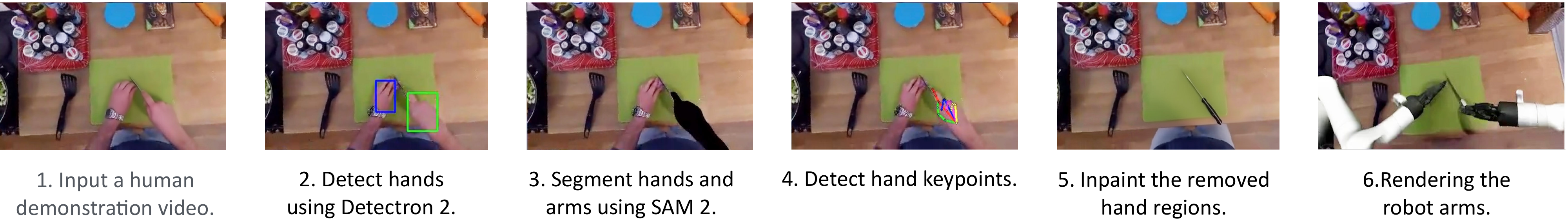}
    \caption{Starting from a human demonstration video, we first detect hands using Detectron2 \cite{wu2019detectron2} and then segment hands and arms using Segment Anything \cite{ravi2024sam}. Next, we perform hand keypoint detection and inpaint the removed hand regions to obtain clean background frames. We then apply inverse kinematics solving to map the detected hand keypoints to robot arm poses and render the robot arms into the videos. Finally, with a human-in-the-loop filtering process, we curate over 6,000 high-quality synthetic human–robot paired videos to support the training of our Mitty model.}
    \label{fig:results}
\end{figure*}

\subsection{Dataset Construction}
\label{sec:dataset_construction}

A key bottleneck in robotic learning lies in data acquisition: collecting real-world robot manipulation data is costly and slow, which limits generalization across large-scale tasks. Meanwhile, ego-centric human activity datasets such as EPIC-Kitchens \cite{damen2020epic}, Ego4D\cite{grauman2022ego4d}, and EgoExo4D\cite{grauman2024ego} have accumulated millions of high-quality demonstrations covering diverse actions and environments. Effectively transferring these large-scale human videos into robotic learning is critical to overcoming the current data bottleneck.

To alleviate the scarcity of human–robot paired videos, we build upon the data rendering approach proposed in the Masquerade\cite{lepert2025masquerade} paper and introduce an automated pipeline. This pipeline takes egocentric human videos as input and produces robot-arm rendered results through the following steps.

\textbf{Hand Pose Estimation:} We use models such as HaMeR for 3D Hand Mesh Recovery to extract 3D hand keypoints and motion trajectories from ego-centric videos.

\textbf{Hand Segmentation and Removal:} We first use Detectron2 \cite{wu2019detectron2} to detect human hands, and apply Segment Anything 2 (SAM2) \cite{ravi2024sam} to perform fine-grained segmentation and remove the hands and forearms from the video.

\textbf{Video Inpainting:} We apply E2FGVI\cite{li2022towards}, a video inpainting model to fill the removed regions across frames, producing clean background videos without hands.

\textbf{Pose Mapping:} The predicted hand keypoints are mapped to robot end-effector poses, including target position (midpoint between thumb and index finger), target orientation (plane normal plus fitted vector), and gripper opening (thresholded thumb–index distance).

\textbf{Robot Arm Rendering:} Using RobotSuite \cite{robosuite2020}, we render robot arms corresponding to the mapped poses into the inpainted videos. Fine-tuning of poses and data cleaning/filtering further improves the fidelity of the resulting paired videos.

Given the multi-step nature of our automated data generation pipeline, cumulative errors and inconsistencies can arise across segments. To mitigate these issues, we employ a human-in-the-loop filtering mechanism to rigorously audit and remove low-quality samples, thereby improving data fidelity and internal consistency. After filtering, each resulting video is further segmented into fixed-length clips sampled at equal intervals, forming the final training and testing sets. This process yields a high-quality human–robot paired dataset that provides strong supervision for In-Context Diffusion Transformer models such as Mitty and establishes a solid foundation for reliable cross-task and cross-environment generalization.

\begin{figure*}[htb]
    \centering
    \includegraphics[width=0.96\linewidth]{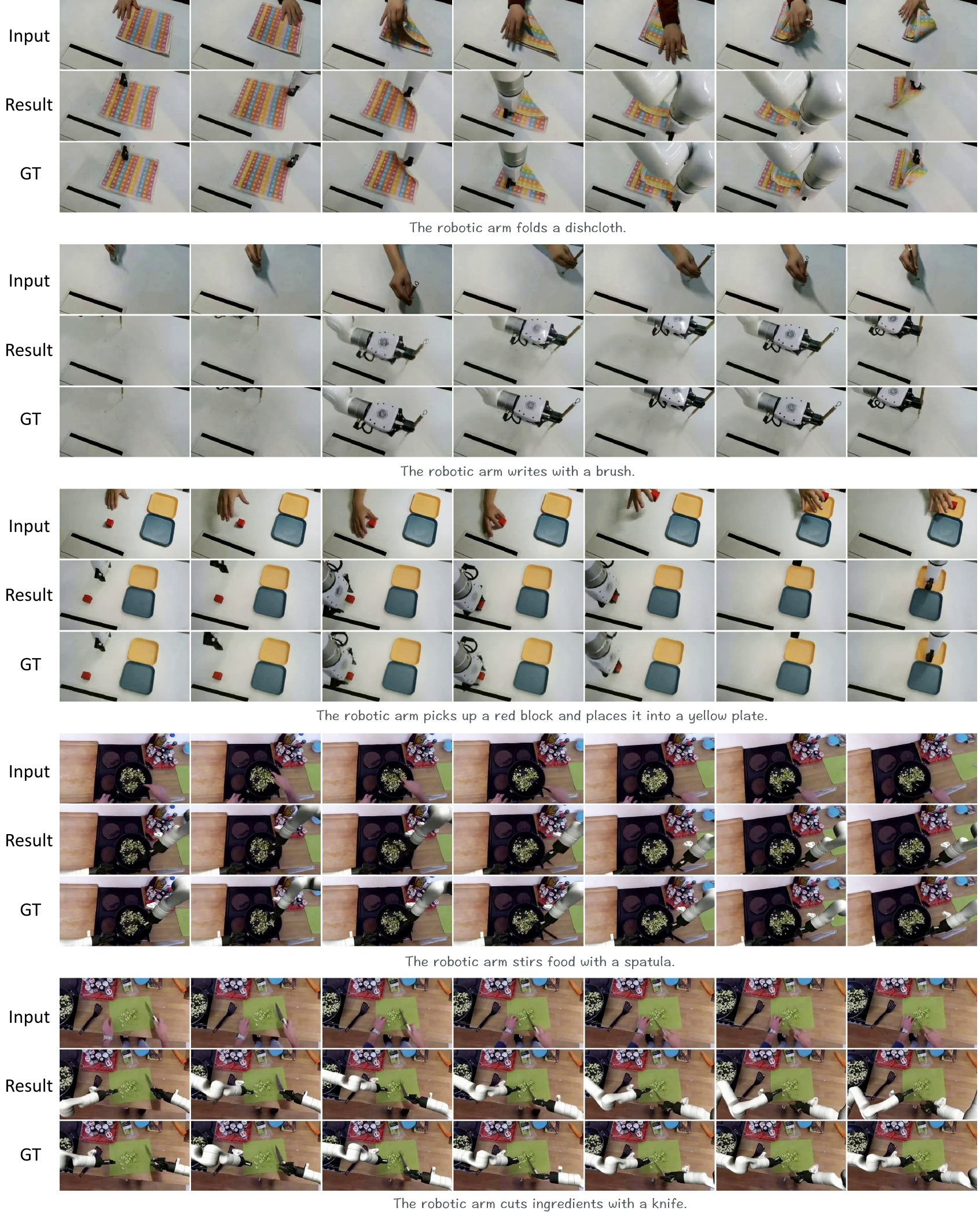}
    \caption{ Mitty’s generation results on Human2Robot and EPIC-Kitchens datasets. In each group of results, the first row shows the human demonstration videos, the second row shows the outputs generated by our method, and the third row shows the ground-truth robot execution videos.}
    \label{fig:results}
\end{figure*}

\section{Experiments }

\subsection{Setup. }


We build on the pretrained Wan 2.2 TI2V-5B dense model and additionally train on the Wan 2.2 TI2V-14B MoE model. We adopt a LoRA-based fine-tuning strategy, simultaneously adapting both high-noise and low-noise branches. The TI2V-5B model is trained for 20k steps, while the larger TI2V-14B model is trained for 10k steps due to computational cost considerations. The LoRA rank is set to 96 with a fixed learning rate of \(1\times10^{-4}\). All experiments are conducted on two H200 GPUs. Both training and inference at a resolution of 416×224, with an effective batch size of 4.

\subsection{Datasets ad Benchmark.}

We evaluate Mitty on two standardized datasets, with all videos resampled to 8 FPS and split into 41-frame clips for consistent temporal resolution. For \textbf{Human2Robot (H2R)}, filtering short or low-quality sequences yields 11,788 paired clips, of which 500 are used for testing. For \textbf{EPIC-KITCHENS}, we apply the synthesis pipeline in Sec.~\ref{sec:dataset_construction} to render robot arms into egocentric videos, resulting in 34,820 clips; 200 form the test set, evenly divided into 100 seen and 100 unseen scenes to evaluate cross-environment generalization. All quantitative results are reported on these held-out splits to ensure fair and consistent benchmarking.

\subsection{Metric. }

We evaluate our model using multiple criteria.
First, video quality is measured using standard metrics, including Fréchet Video Distance (FVD), PSNR, MSE, and SSIM, and is evaluated on the test sets of both datasets. Second, task success rate serves as our primary metric: videos exhibiting obvious visual artifacts, temporal discontinuities or distortions, or incorrect robot-arm motions are considered failures. Three domain experts independently review all generated videos, and disagreements are resolved through discussion to reach consensus.

In addition, for Embodiment Consistency, which evaluates whether the generated robot arm matches the reference arm in visual appearance, we compute a normalized average score using CLIP image score, DreamSim, and human evaluation.

We further measure Human Preference when comparing against baseline methods. All generated videos are anonymized, and for each input, outputs from different methods are presented simultaneously to users, who are asked to select the video with the highest overall quality. We received 20 valid survey responses for this user study.

\subsection{Baseline Methods.}

We compare Mitty against two groups of baselines. The first group consists of models built on the Wan 2.2 family and trained using our proposed in-context learning setup. Our primary configuration uses the 5B model with first-frame conditioning and human reference videos, while a larger 14B model is also evaluated to study scaling. We further include two ablations—first-frame only (removing the human reference video) and text-free (removing the task description)—as well as a comparison between separate training (one model per dataset) and mixed training (joint training across datasets). The second group consists of general-purpose video editing methods, including Aleph~\cite{aleph}, Kling~\cite{kling}, and MoCha~\cite{mocha}, which perform image-conditioned editing; Kling and MoCha additionally require a SAM-generated mask to specify the arm region to replace. Together, these baselines enable a comprehensive evaluation across both diffusion-based in-context learning models and generic video editing pipelines.

\subsection{Results. }
\textbf{Quantitative Evaluation} Figure~\ref{fig:results} shows Mitty’s qualitative results on Human2Robot and EPIC-Kitchens.
Each group contains three rows: the first row is the human demonstration, the second row is Mitty’s zero-frame generation without first-frame conditioning, and the third row is the ground truth robot-execution video.
We observe that Mitty accurately preserves scene layout and object interactions while producing smooth, temporally coherent robot motions. Mitty also generalizes robustly to unseen tasks and environments, maintaining strong visual consistency, action consistency, and background stability.

\noindent \textbf{Qualitative Results} Table~\ref{tab1} summarizes our results on the Human2Robot and EPIC-Kitchens datasets. Across both datasets, adding the first-frame condition consistently reduces FVD and MSE while slightly increasing PSNR, SSIM, and SR, demonstrating more stable and faithful video generation. On Human2Robot, our larger T2V 14B model achieves the best overall performance, yielding the lowest FVD and MSE and the highest PSNR, SSIM, and SR compared to TI2V 5B. In contrast, the EPIC-Kitchens dataset presents more diverse scenes, more complex environments, and moving camera viewpoints, which make the task significantly more challenging. Consequently, performance metrics on EPIC-Kitchens are generally lower than on Human2Robot, reflecting the increased difficulty of achieving high-fidelity generation under such conditions.

Table \ref{tab-f} compares our method with baseline approaches across task-level success rate (SR), human preference, and embodiment consistency. As a rendering-based pipeline, Masquerade achieves the highest embodiment consistency by a large margin, since the rendered robot arm is directly composited into the scene. However, its task success rate is significantly lower due to the heavy error accumulation across multiple stages. In contrast, general video editing methods (Kling, Mocha, Aleph) exhibit poor embodiment consistency, as a single reference image is insufficient to maintain stable robotic appearance and structure throughout the sequence. Our method achieves the best task success rate and human preference, as well as the second-highest embodiment consistency, demonstrating a strong balance between correctness, visual fidelity, and structural stability.

\begin{table}[htb]
\footnotesize
\setlength{\tabcolsep}{2.2pt}
\renewcommand{\arraystretch}{0.9}
\centering
\hspace*{-2mm}
\caption{Video generation metrics on Human2Robot and EPIC-Kitchens datasets. 
Lower FVD/MSE and higher PSNR/SSIM/SR (\%) indicate better quality. 
\textit{w/o 1st f} denotes generation without conditioning on the first frame, 
while \textit{w 1st f} denotes conditioning on it. 
EPIC-Kitchens results are divided into \textit{Seen} and \textit{Unseen} environments. The best results are highlighted in bold.}
\label{tab1}
\begin{tabular}{>{\centering\arraybackslash}m{1.4cm} l ccccc}
\toprule
\textbf{Dataset} & \textbf{Meth./Set.} & \textbf{FVD$\downarrow$} & \textbf{PSNR$\uparrow$} & \textbf{SSIM$\uparrow$} & \textbf{MSE$\downarrow$} & \textbf{SR$\uparrow$} \\
\midrule
\multirow{3}{*}{\makecell[c]{Human\\2Robot}}
 & TI2V 5B (w/o 1st f)   & 7.96 & 21.5 & 0.835 & 0.0084 & 85 \\
 & TI2V 5B (w 1st f)     & 7.40 & 21.7 & 0.837 & 0.0081 & 91 \\
 & T2V 14B               & \textbf{6.48} & \textbf{22.7} & \textbf{0.851} & \textbf{0.0069} & \textbf{93} \\
\midrule
\multirow{3}{*}{\makecell[c]{EPIC-\\Kitchens\\(Seen)}}
 & TI2V 5B (w/o 1st f)  & 7.65 & 13.40 & 0.630 & 0.0508 & 84 \\
 & TI2V 5B (w 1st f)    & 7.23 & 13.46 & 0.617 & 0.0494 & 88 \\
 & T2V 14B              & \textbf{6.90} & \textbf{13.69} & \textbf{0.634} & \textbf{0.0466} & \textbf{90} \\
\midrule
\multirow{3}{*}{\makecell[c]{EPIC-\\Kitchens\\(Unseen)}}
 & TI2V 5B (w/o 1st f) & 9.74 & 13.30 & 0.670 & 0.0495 & 79 \\
- & TI2V 5B (w 1st f)   & 9.48 & 13.29 & 0.627 & 0.0493 & 86 \\
 & T2V 14B             & \textbf{9.35} & \textbf{13.32} & \textbf{0.673} & \textbf{0.0479} & \textbf{89} \\
\bottomrule
\end{tabular}
\end{table}

\begin{table}[htb]
\footnotesize
\setlength{\tabcolsep}{2.2pt} 
\renewcommand{\arraystretch}{0.9} 
\centering
\hspace*{-2mm} 
\caption{Ablation study on Human2Robot and EPIC-Kitchens datasets under three settings: 
(1) \textit{w/o ref vid.} (without reference video), 
(2) \textit{w/o task desc.} (without task description), and 
(3) \textit{Full model} with separate or mixed training. 
EPIC-Kitchens results are divided into \textit{Seen} and \textit{Unseen} environments. 
Lower FVD/MSE and higher PSNR/SSIM/SR (\%) indicate better quality. 
The best results are highlighted in bold.}
\label{tab:ablation} 
\begin{tabular}{>{\centering\arraybackslash}m{1.4cm} l ccccc}
\toprule
\textbf{Dataset} & \textbf{Meth./Set.} & \textbf{FVD$\downarrow$} & \textbf{PSNR$\uparrow$} & \textbf{SSIM$\uparrow$} & \textbf{MSE$\downarrow$} & \textbf{SR$\uparrow$} \\
\midrule
\multirow{4}{*}{\makecell[c]{Human\\2Robot}}
 & w/o ref vid.        & 9.43 & 20.05 & 0.818 & 0.0091 & 65 \\
 & w/o task desc.      & 8.42 & 21.42 & 0.837 & 0.0091 & 88 \\
 & Full (Mixed train.) & 9.54 & 16.63 & 0.742 & 0.0138 & 72 \\
 & Full (Sep. train.)  & \textbf{7.40} & \textbf{21.7} & \textbf{0.837} & \textbf{0.0081} & \textbf{91}  \\
\midrule
\multirow{4}{*}{\makecell[c]{EPIC\\Kitchens\\(Seen)}}
 & w/o ref vid.        & 12.25 & 12.22 & 0.534 & 0.0728 & 75 \\
 & w/o task desc.      & 9.43 & 13.05 & 0.602 & 0.0508 & 83 \\
 & Full (Mixed train.) & 8.31 & 13.39 & \textbf{0.617} & 0.0499 & 81 \\
 & Full (Sep. train.)  & \textbf{7.23} & \textbf{13.46} &\textbf{0.617} & \textbf{0.0494} & \textbf{88} \\
\midrule
\multirow{4}{*}{\makecell[c]{EPIC\\Kitchens\\(Unseen)}}
 & w/o ref vid.        & 10.31 & 12.65 & 0.531 & 0.0734 & 71 \\
 & w/o task desc.      & 9.82 & 12.92 & 0.597 & 0.0526 & 82 \\
 & Full (Mixed train.) & 9.73 & \textbf{13.75} & 0.613 & \textbf{0.0463} & \textbf{86} \\
 & Full (Sep. train.)  & \textbf{9.48} & 13.29 & \textbf{0.627} & 0.0493 & 81 \\
\bottomrule
\end{tabular}
\end{table}

\begin{table}[t]
\centering
\footnotesize
\caption{Comparison of our method and baselines in task-level SR, human preference, and embodiment consistency.}
\label{tab-f}
\begin{tabular}{lccc}
\toprule
\textbf{Method} 
& \textbf{\makecell{Task-level\\SR (\%)}} 
& \textbf{\makecell{Human\\Preference (\%)}} 
& \textbf{\makecell{Embodiment\\Consistency}} \\
\midrule
Masquerade & 31.5 & 20.0 & \textbf{96.5} \\
Kling      & 70.0 & 4.8  & 77.4 \\
Mocha      & 69.0 & 4.0  & 60.2 \\
Aleph      & 78.0 & 3.2  & 73.9 \\
Ours       & \textbf{84.5} & \textbf{68.0} & 92.6 \\
\bottomrule
\end{tabular}
\end{table}

\subsection{Ablation Study.}

Table~\ref{tab:ablation} presents the ablation results on the Human2Robot and EPIC-Kitchens datasets using the TI2V-5B model. Considering the additional training and inference cost, we adopt the TI2V-5B with first-frame conditioning as our default baseline. When the human reference video is removed, the model predicts subsequent frames using only the initial robot frame and task description, resulting in clear degradation in FVD, PSNR, SSIM, and SR on both datasets. In contrast, removing task description prompts causes only minor changes, indicating that Mitty relies more on visual demonstrations than textual cues. Notably, the impact of removing the human reference video is more severe on EPIC-Kitchens due to its more diverse scenes and moving camera viewpoints, further emphasizing the importance of strong visual conditioning under complex environments. Finally, because the two datasets differ substantially in tasks and environments (e.g., single-arm vs. dual-arm manipulation and varying scene complexity), the full model trained separately on each dataset outperforms mixed training.

\begin{figure}[htbp!]
    \centering
    \includegraphics[width=1.0\linewidth]{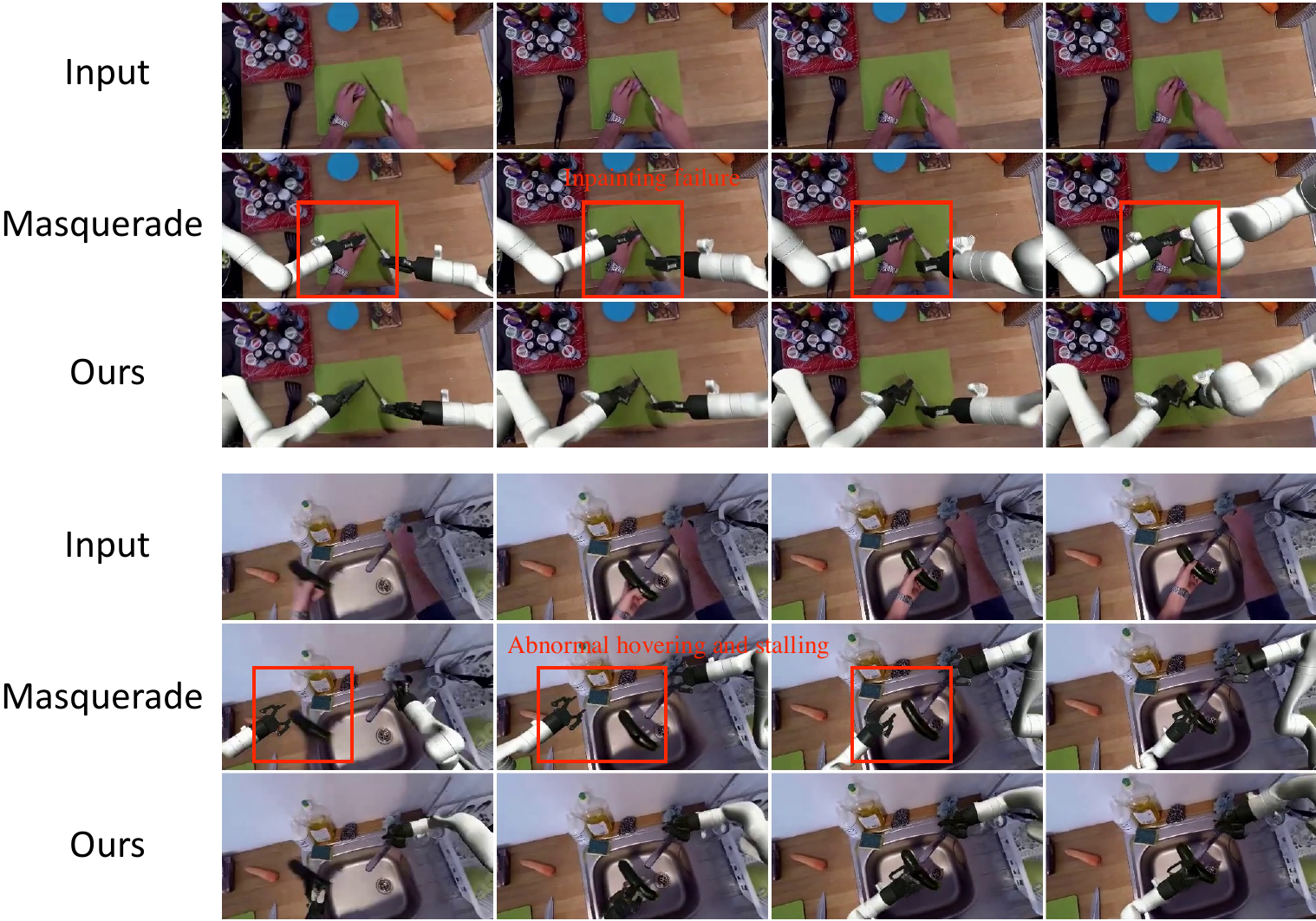}
    \caption{Masquerade’s multi-stage pipeline is prone to compounded errors (e.g., joint detection, inpainting, and rendering failures), as highlighted in red. In contrast, our curated training data enables a robust end-to-end model that produces more reliable Human2Robot mappings.}
    \label{compare1}
\end{figure}

\subsection{Comparison with Masquerade’s Data Pipeline.}

Figure \ref{compare1} provides a qualitative comparison between our method and Masquerade. Masquerade employs a multi-stage rendering pipeline—including hand segmentation, keypoint estimation, background inpainting, and robot-arm rendering—which is prone to error accumulation and often produces noticeable artifacts. Typical issues include unstable hand masks, drifting or missing keypoints, residual artifacts from incomplete inpainting, and misaligned or physically implausible robot-arm renderings such as penetration or floating. These errors frequently compound across stages and degrade both visual quality and physical realism. We present additional representative failure modes of Masquerade in the supplementary materials.

Importantly, our comparison is not aimed at competing with the full Masquerade system, but rather at analyzing the multi-stage, rendering-based data synthesis pipeline that Masquerade represents. Although such pipelines suffer from structural limitations and a relatively low usable-data rate, generating large-scale raw data with Masquerade and applying human-in-the-loop filtering allows us to curate a high-quality paired dataset. This curated subset is sufficient to train a robust and strongly generalizable end-to-end Human2Robot video generation model.

\subsection{Compare with Video Editing Methods.}

\begin{figure}[t!]
    \centering
    \includegraphics[width=1.0\linewidth]{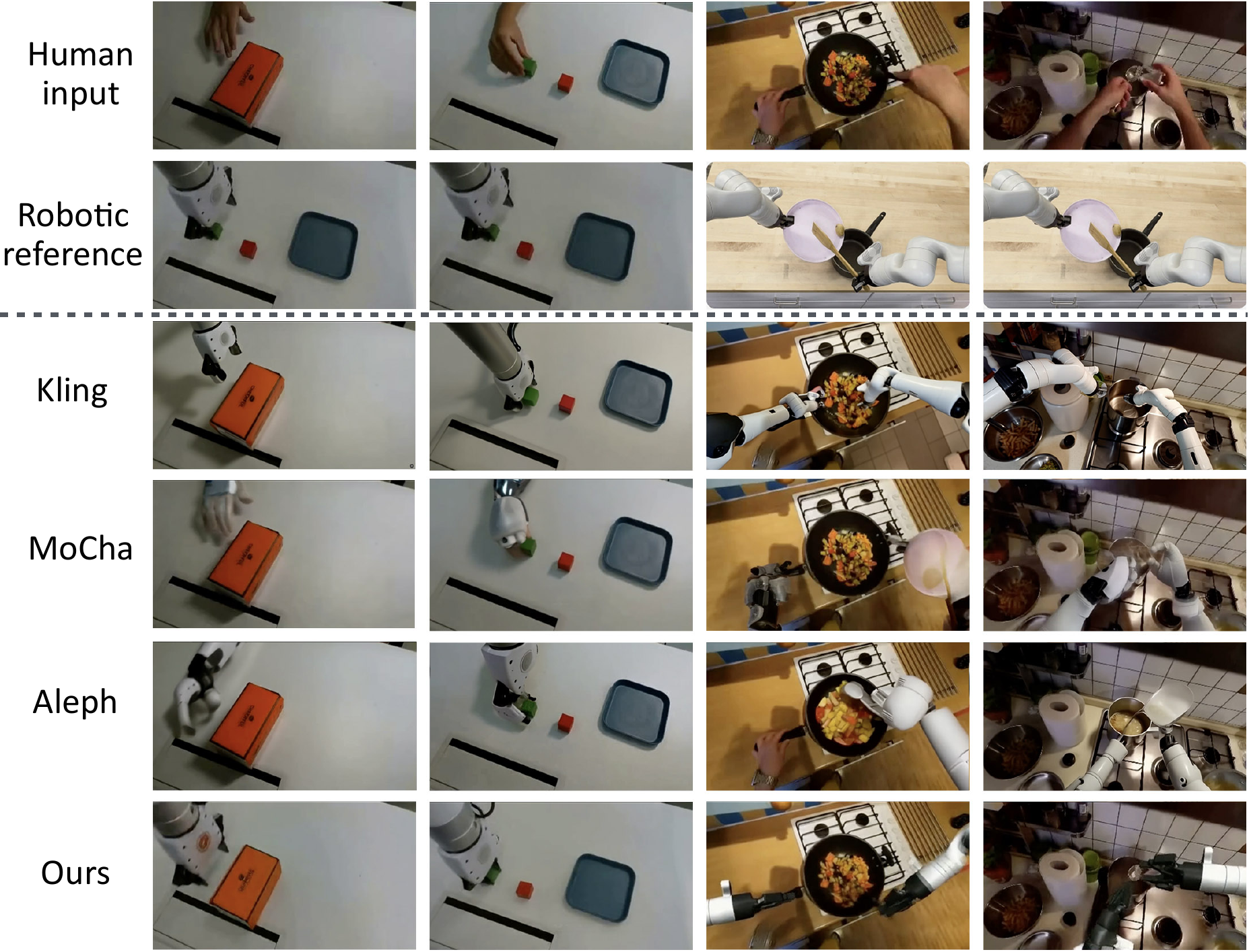}
    \caption{Compared with state-of-the-art video editing models, the baseline methods take a robot reference image and a human demonstration video as input. However, even the most advanced baselines still struggle to maintain appearance and structural consistency of the robotic arm throughout the sequence.}
    \label{compare2}
\end{figure}

Figure \ref{compare2} compares our method with three state-of-the-art video editing approaches (two commercial APIs and one open-source method). These methods take the input video and a single robot-arm reference image, using prompts to replace the human arm with the robot arm. However, the results show that a single reference image is insufficient to define the robot arm’s appearance and structure, leading even the most advanced models to produce deformation, structural errors, and distortions. In contrast, our paired-data–trained model consistently preserves the correct appearance and structure. This highlights that Human2Robot remains a challenging task that cannot be solved by generic video editing models and requires sustained, dedicated research efforts.

\section{Limitations and Future Work}

While Mitty demonstrates strong performance in video-level action generation and cross-task generalization, it is not yet a complete Video Policy pipeline. The current model can generate robot-arm execution videos but cannot explicitly output control-ready action sequences, and thus has not been evaluated in a full closed-loop setting on real robots. In addition, task success rate is still based on expert assessments of generated videos rather than physical rollouts.

Nevertheless, high-fidelity and semantically aligned Human2Robot video generation provides an essential foundation for enabling future video-to-policy inversion, offering structured supervisory signals that are difficult to obtain from sparse representations alone. Moving forward, we plan to incorporate action or policy prediction into the framework, conduct closed-loop experiments in simulation and on real hardware, and develop more automated and physically grounded evaluation metrics, ultimately advancing the Human2Robot task toward a complete Video Policy solution.
 
\section{Conclusion}

In this work, we introduced Mitty, a Diffusion Transformer that enables end-to-end Human2Robot video generation through in-context learning. Built upon the Wan 2.2 backbone and a scalable paired-data synthesis pipeline, Mitty bypasses keypoints and trajectories to directly translate human demonstrations into temporally aligned robot-execution videos. Experiments on Human2Robot and EPIC-Kitchens show strong performance, solid generalization, and clear advantages over rendering pipelines and video editing systems. Ablation studies further highlight the effectiveness of in-context conditioning and mixed training. Mitty provides a meaningful starting point for future video-to-policy research and real-world robot learning, and opens up opportunities to explore more complex tasks and tighter human–robot mapping in the future.

\newpage

{
    \small
    \bibliographystyle{ieeenat_fullname}
    \bibliography{main}
}

\clearpage
\appendix
\input{sec/X_suppl}

\end{document}

%% file: sec/X_suppl.tex
\clearpage
\setcounter{page}{1}
\maketitlesupplementary

\setcounter{section}{0}
\renewcommand{\thesection}{\Alph{section}}

\section{Evaluation Protocol and Details}

This section provides detailed descriptions of our quantitative metrics, qualitative assessments, and the expert evaluation protocol.

\subsection{Evaluation Metric}

We evaluate our model using the following metrics:
\begin{itemize}
\item \textbf{FVD}: overall video quality and temporal coherence.
\item \textbf{PSNR}, \textbf{SSIM}, \textbf{MSE}: frame-level fidelity and structural consistency.
\item \textbf{Task Success Rate (SR)}: evaluated as part of the human expert assessment.
\end{itemize}

\paragraph{Definition of Task Success.}
A clip is considered successful when the robot correctly follows the human reference video and completes the intended task.
Minor artifacts—such as small geometric distortions, local glitches, or brief generative inconsistencies—are ignored as long as they do not affect task execution.
A clip is considered a failure when clear issues such as action mismatches, implausible interactions, hovering or drifting behavior, or any form of task incompletion are observed.

\subsection{Human Expert Evaluation}

Task Success Rate is derived from a structured human evaluation process.
All generated videos are anonymized and randomly shuffled.
Three domain experts independently assign a binary judgment (\emph{success} or \emph{failure}) for each clip.
For cases with inconsistent labels, the experts discuss until a consensus is reached.

This protocol ensures consistent and reliable qualitative evaluation.

\section{Masquerade Typical Failure Cases}

In Fig.~1 of the main paper, where Masquerade is still able to render frames normally, we illustrate several representative failure types. These error cases have already been described in the main text.

However, Masquerade frequently fails even earlier at the rendering stage, producing multiple consecutive all-black outputs. We further analyze the causes of these rendering failures and categorize them into the following three types.

\begin{figure*}[htb]
    \centering
    \includegraphics[width=1.0\linewidth]{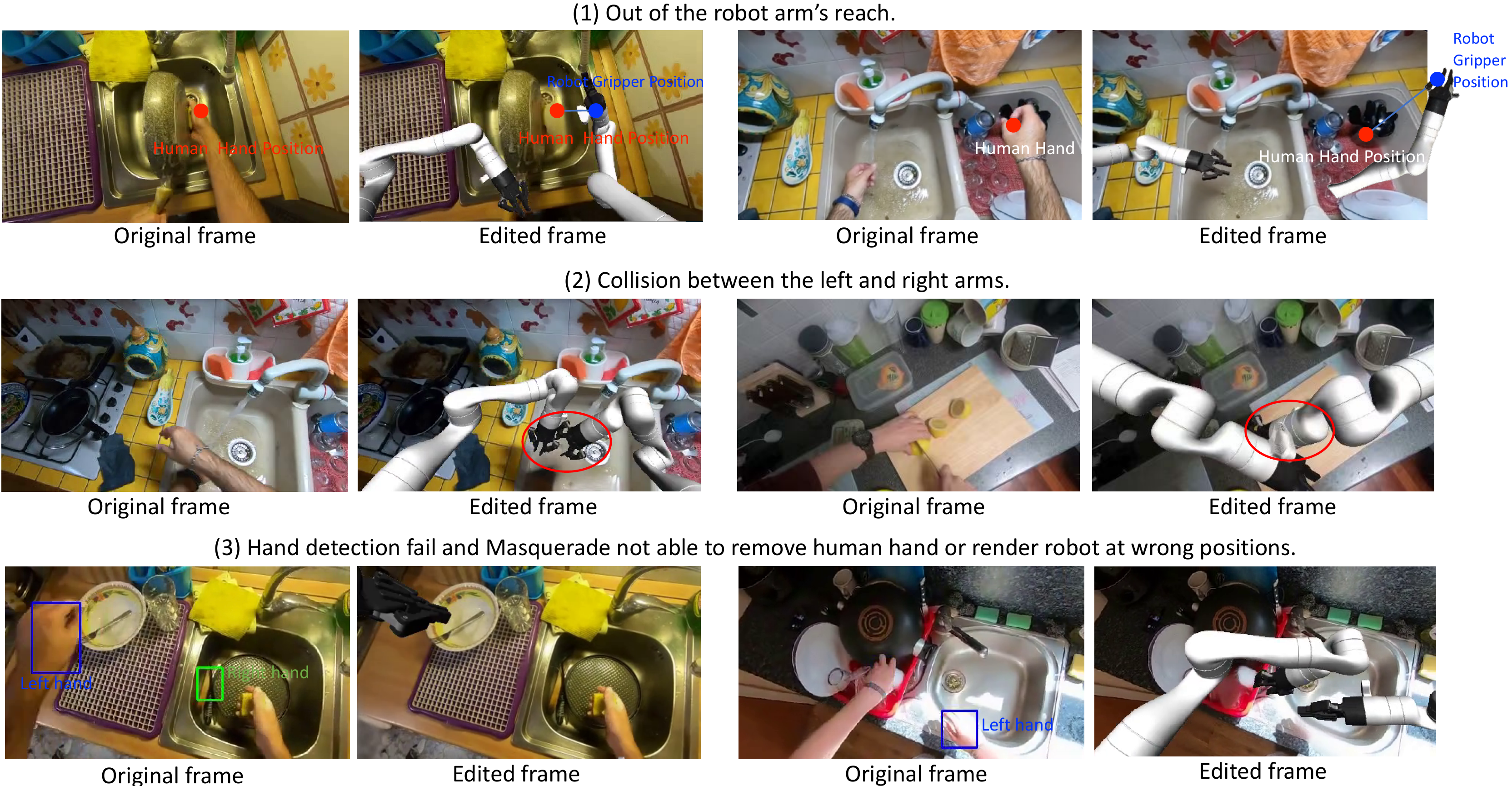}
    \caption{
    Representative failure cases of Masquerade.
From top to bottom:
(1) Out-of-reach failure — when the human hand appears outside the robot’s feasible workspace, Masquerade renders the robot gripper at physically unreachable positions, causing the arm to float above the scene rather than interacting with objects.
(2) Collision between the left and right arms — incorrect estimation of the spatial relationship between the two hands leads to intersecting or colliding robot-arm trajectories, resulting in unrealistic bi-manual interactions.
(3) Hand detection failure and incorrect rendering — Masquerade fails to correctly detect or remove the human hand, producing inaccurate left/right hand boxes or residual artifacts, and often rendering the robot arm in wrong or inconsistent positions.
    }
    \label{fig:failure_cases}
\end{figure*}

\noindent\textbf{Out of the Robot Arm’s Reach.}
The First failure case arises when the human hand appears outside the robot arm’s feasible workspace. Since Masquerade does not explicitly account for kinematic reachability or workspace limits, the system attempts to place the robot gripper at locations the robot cannot physically reach. As shown in Fig.~\ref{fig:failure_cases}  (1), the rendered robot arm floats in the air or points toward an unreachable region, leading to failed interaction with the target object and breaking realism.

\noindent\textbf{Collision Between the Left and Right Arms.}
The second failure mode occurs when Masquerade incorrectly estimates the spatial relationship between the left and right hands, causing the corresponding robot arms to collide in 3D space. As shown in Fig.~\ref{fig:failure_cases}  (2), the generated trajectories of the two arms intersect, producing unrealistic physical interactions and invalid manipulation behavior. These self-collisions highlight Masquerade’s inability to model inter-arm spatial constraints and coordinated bi-manual motion.

\noindent\textbf{Hand Position Error.}
Reproducing Masquerade’s hand-detection pipeline reveals that its single-arm Detectron model \cite{wu2019detectron2} cannot reliably distinguish between left and right hands. We use ViT-Pose \cite{xu2022vitpose}, a state-of-the-art whole-body pose estimator. Compared to Detectron2 and other pose estimators, it provides more reliable left/right hand recognition since its detailed hand keypoints are predicted for each hand separately, allowing us to distinguish hands better.

However, even with this improvement, Masquerade still frequently exhibits hand-position errors. As illustrated in Fig.~\ref{fig:failure_cases} (3), inaccurate hand localization causes the predicted hand regions to drift away from the true human hand position. Consequently, the system renders the robot gripper at an incorrect location—often on the opposite side of the scene—leading to immediate task failure because the robot attempts to act on objects at a physically incorrect or unreachable location.

 Besides, Masquerade cannot reliably handle single-hand scenarios. In Epic-kitchen, the method assumes a custom bimanual setup composed of two independent single-arm robots. When one hand is not detected, it falls back to a predefined neutral pose, which can lead to incorrect or failed rendering. Although hiding the undetected arm may appear to be a straightforward alternative, the model lacks the ability to determine whether the absence of a hand is actually part of the scene semantics. For example, the user may be genuinely using only one hand, or the missing hand may simply result from a transient failure of the hand detector. Without this distinction, blindly hiding the arm could incorrectly remove a hand that should still be present, leading to inconsistent or erroneous robot rendering. Therefore, we retain the original configuration to remain consistent with the method’s intended design.

\section{Additional Experimental Results}

\begin{figure*}[htb]
    \centering
    \includegraphics[width=1.0\linewidth]{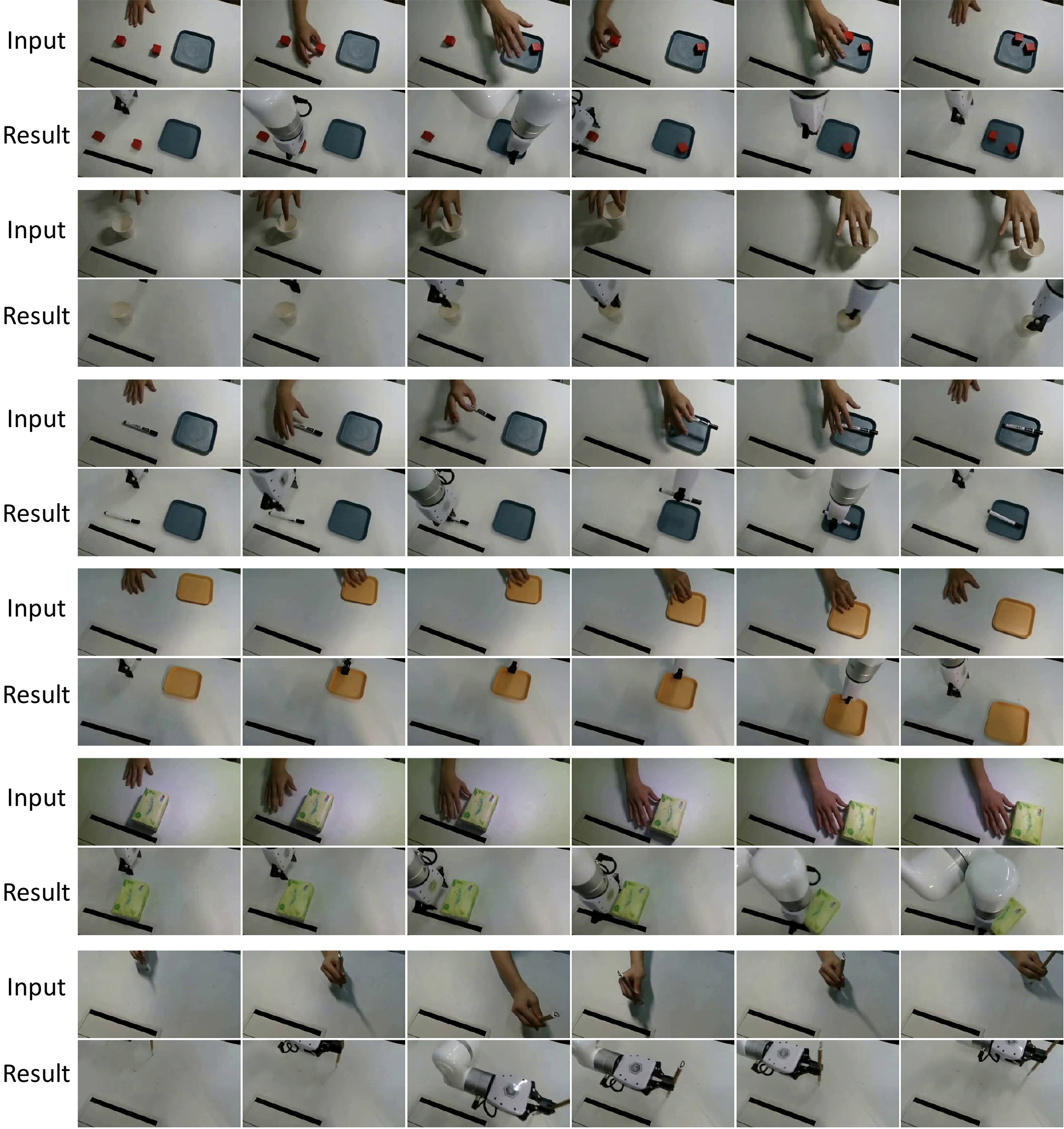}
    \caption{ More Mitty’s generation results on Human2Robot datasets. In each group of results, the first row shows the human demonstration videos, the second row shows the outputs generated by our method.}
    \label{fig:SM1}
\end{figure*}

\textbf{Human2Robot Results.}
We provide more qualitative results on Human2Robot tasks in Fig.~\ref{fig:SM1} , including:
grasping, placing, cooking-related actions, countertop manipulation, and basic tool usage.  
Our model consistently maintains scene structure, motion coherence, and identity stability.

\begin{figure*}[htb]
    \centering
    \includegraphics[width=1.0\linewidth]{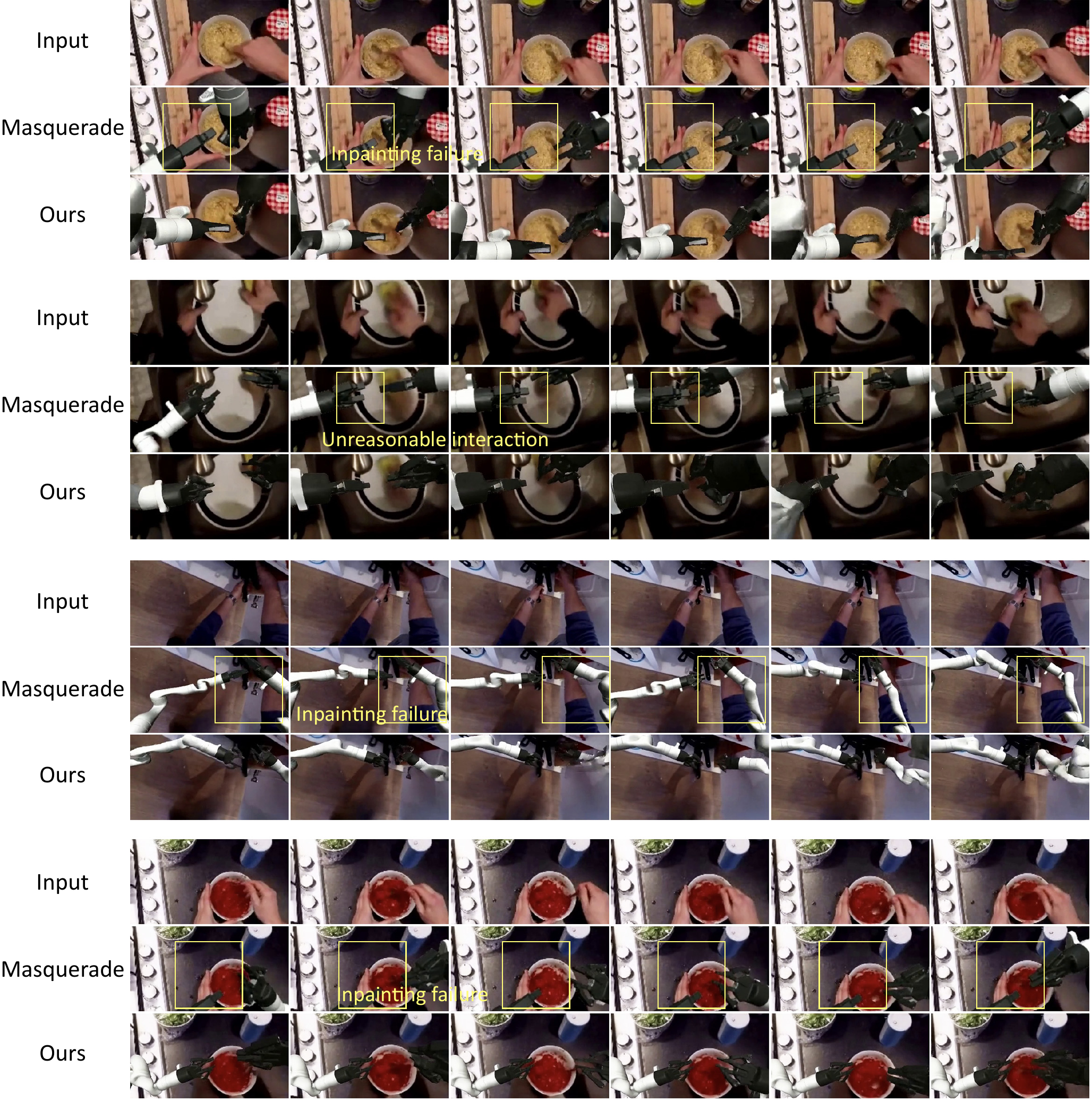}
    \caption{ More comparative experimental results. In each group, the first row shows the input human demonstration video, the second row presents the results produced by Masquerade, and the third row displays the results generated by our method.}
    \label{fig:SM2}
\end{figure*}

\noindent\textbf{Comparison with Masquerade on EPIC-Kitchens Dataset.}
In Fig.~\ref{fig:SM2}, on the EPIC-Kitchens dataset, we present additional examples from both seen and unseen kitchen environments, demonstrating robustness to camera shake, complex scenes, lighting variations, and multi-object interactions. Side-by-side comparisons with Masquerade further highlight its typical failure patterns—including hovering artifacts, black-frame outputs, arm distortion, inpainting errors, and incorrect gripper orientations—showing that our method achieves more stable and consistent results under the same conditions.

\section{Our Failure Case Analysis}

\begin{figure*}[htb]
    \centering
    \includegraphics[width=1.0\linewidth]{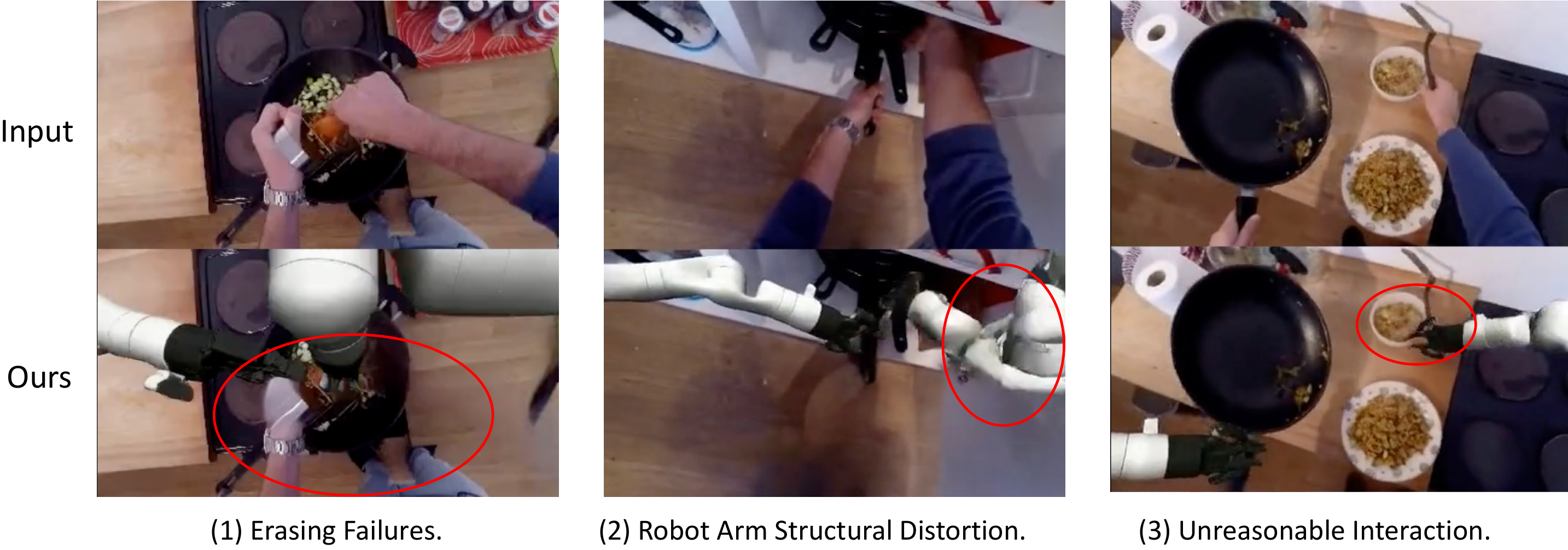}
    \caption{The model sometimes produces structural distortions, such as unnatural joint angles or anatomically implausible robot-arm shapes—an error mode particularly common on EPIC-Kitchens due to complex hand–object interactions. It may also exhibit unreasonable interactions, where the robot arm fails to follow physically plausible trajectories, misses or drifts past the target object, or interacts with nonexistent items.}
    \label{fig:SM3}
\end{figure*}

In addition to the results presented in the main paper, Fig.~\ref{fig:SM3} illustrates several representative failure cases of our GPT-based approach.
We categorize these failures into three major types:

\noindent\textbf{Erasing Failures.}
Regions that should have been replaced or removed are not fully erased, leading to incomplete transitions or visual remnants from the source video.

\noindent\textbf{Robot Arm Structural Distortion.}
The generated robot arm exhibits geometric inconsistencies, unnatural joint angles, or anatomically impossible shapes.
Notably, this is the most frequent failure mode on the EPIC-Kitchens dataset, likely due to complex hand–object interactions and challenging first-person motion patterns.

\noindent\textbf{Unreasonable Interaction.}
The robot arm’s motion does not follow physically plausible trajectories or fails to maintain correct contact with manipulated objects. Examples include missing the target, drifting past the object, or interacting with nonexistent items.

